\documentclass[11pt,a4paper]{article}

\usepackage[margin=25mm]{geometry}
\usepackage{amsmath,amssymb}
\usepackage{graphicx}
\usepackage{fontawesome5}
\usepackage{booktabs}
\usepackage[super,sort&compress]{natbib}
\usepackage{lineno}
\usepackage[hidelinks]{hyperref}

\hypersetup{pdftitle={A verifiable foundation for automated computational science}}

\usepackage{xcolor}

\title{FARMS: A Verifiable Foundation for Automated \\Computational Science}
\author{%
  Di Wu$^{1}$, Ling Liang$^{2}$ and Haizhao Yang$^{1,3,*}$\\[6pt]
  \parbox{0.94\textwidth}{\centering\small
    $^{1}$Department of Mathematics, University of Maryland,
    College Park, USA\\
    $^{2}$Department of Mathematics, The University of Tennessee,
    Knoxville, USA\\
    $^{3}$Department of Computer Science, University of Maryland,
    College Park, USA\\[4pt]
    $^{*}$Corresponding author:
    \href{mailto:hzyang@umd.edu}{hzyang@umd.edu}
  }
}
\date{}

\begin{document}

\maketitle

\begin{abstract}
Automated computational science could accelerate discovery by selecting, adapting and developing numerical methods. This prospect requires mathematical knowledge that AI systems can interpret and reuse with explicit guarantees. Here we present FARMS, a faithful and efficient foundation that transforms computational mathematics into AI-ready computing knowledge. FARMS connects formal statements to source context, algorithmic meaning and quantitative guarantees through source preparation, reusable representations and semantic review. Across 171 targets in randomized numerical linear algebra and numerical partial differential equations, 157 were accepted by the pipeline, with mean estimated costs of US \$0.85–2.29 per target across sources. Lean-checked implications from explicit source premises connect selected statements to computational skills, illustrated by a heat-flow solve and wave-response compression. FARMS therefore supports an affordable, reusable knowledge foundation for trustworthy automated computational science.
\end{abstract}

\begin{center}
  \small
  \faGithub\enspace\textbf{GitHub repo:}\enspace
  \url{https://github.com/Woody2357/FARMS-Computational-Science}\\[0.3em]
  \raisebox{-0.15em}{\includegraphics[height=1em]{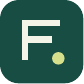}}\enspace
  \textbf{Project page:}\enspace
  \url{https://woody2357.github.io/FARMS-Computational-Science/}
\end{center}


\section{Introduction}

Artificial intelligence (AI) is expanding the ways in which scientific knowledge can be discovered and used \cite{wang2023scientific, boiko2023autonomous,davies2021advancing}. One opportunity is to develop the computational methods that make new investigations possible. Advances in automated algorithm discovery \cite{novikov2025alphaevolve}, from matrix multiplication \cite{fawzi2022discovering} to program search \cite{romera2024mathematical}, show that AI can contribute to the machinery of computation itself. Extending this capability across computational science calls for a mathematical foundation that AI systems can use to select methods, combine their guarantees and reason about proposed improvements. Such a foundation would allow automated discovery to build on the accumulated knowledge of computational mathematics, with the conditions supporting each computational decision available for inspection and checking.

Randomized numerical linear algebra and numerical partial differential equations (PDEs) are central to this opportunity. Randomized methods make large matrix problems tractable through sampling, sketching and low-rank approximation, supporting efficient analysis and computation at scale \cite{martinsson2020randomized,halko2011finding}. Numerical PDE methods connect mathematical models of physical systems to simulations through discretization, stability and error analysis \cite{larsson2003partial}. Randomized sketching also yields preconditioners that accelerate the iterative solution of large least-squares and linear systems \cite{rokhlin2008fast,meng2014lsrn}, while low-rank representations compress families of simulated responses for repeated use \cite{benner2015survey}, connecting linear algebra to the computation and reuse of simulation results. Making their mathematical knowledge accessible to AI would provide reusable components for developing computational methods across a broad range of scientific problems.

Much of this knowledge is expressed through relationships among theorems, algorithms and surrounding explanations. A randomized approximation guarantee depends on how samples are drawn, which outputs the algorithm can construct and how success is measured. A PDE estimate may depend on boundary conditions, regularity and a particular error norm. These relationships determine whether a guarantee can support a computation. An AI-ready representation must therefore be faithful to the computational meaning of the source, preserving the conditions that connect a method to its guarantees. Constructing such representations must also be efficient if researchers are to extend them across the growing literature and share them as a community resource.

Formal mathematics provides a foundation for explicit representations and machine-checked reasoning \cite{avigad2024mathematics,moura2021lean,mathlib2020,hales2017formal}. Advances in neural theorem proving \cite{polu2020generative,trinh2024solving,ren2025deepseek,hubert2026olympiad} and autoformalization \cite{wu2022autoformalization,jiang2023draft} have broadened access to this foundation, supported by shared mathematical benchmarks \cite{zheng2022minif2f,azerbayev2023proofnet}. These efforts increasingly encompass textbook libraries and applied mathematics \cite{gloeckle2026automatic,rammal2026formalizing,long2026cambench,yang2026construction}. Recent work in numerical analysis  \cite{meek2026formalizing} and physics reasoning \cite{meadows2026formalscience} report that compilation-based metrics overstate formalization quality and audits translations for semantic correctness and library reuse. Research on statement evaluation draws the same distinction between successful compilation and faithful representation of a source claim \cite{poiroux2025reliable,ilin2026sorries}. For computational science, these developments create an opportunity to connect mathematical literature with the practical reasoning needed to use numerical methods. This requires representations that retain algorithmic operations, quantitative guarantees and the dependencies needed to combine them in scientific workflows.

Here we present \textbf{FARMS} (Faithful Alignment and Representation of Mathematical Statements), a faithful and efficient foundation for automated computational science. FARMS turns mathematical literature into AI-ready computing knowledge by connecting formal statements to their source context and computational interpretation. Source preparation recovers the algorithms and definitions needed to interpret each claim; reusable domain representations support formalization; and compilation, semantic review and bounded repair assess the resulting statements. Selected statements can then be combined through Lean-checked implications to form computational skills with explicit applicability conditions. This organization connects knowledge construction to its subsequent use in numerical reasoning and computation.

We study 171 targets from four sources in randomized numerical linear algebra and numerical PDEs \cite{martinsson2020randomized,halko2011finding,larsson2003partial,drineas2017lectures}. The cumulative collection contains 157 pipeline-accepted formalizations, with mean estimated costs of US \$0.85–2.29 per target across sources. On six randomly selected targets, FARMS costs an estimated US \$2.84 per target, compared with US \$58.02 for our statement-level adaptation of AutoformBot. We further demonstrate knowledge reuse in a steady heat-flow solve and compression of wave-response libraries, where mathematical guarantees guide accuracy decisions. Together, these results support a practical route to a verifiable foundation for automated computational science: faithful computing knowledge, checked connections to applications and efficient construction that can lower the cost of community participation in trustworthy automated scientific discovery.

\section{Results}
\subsection{Building AI-ready computing knowledge}

\begin{figure}[p]
\centering
\includegraphics[width=\linewidth]{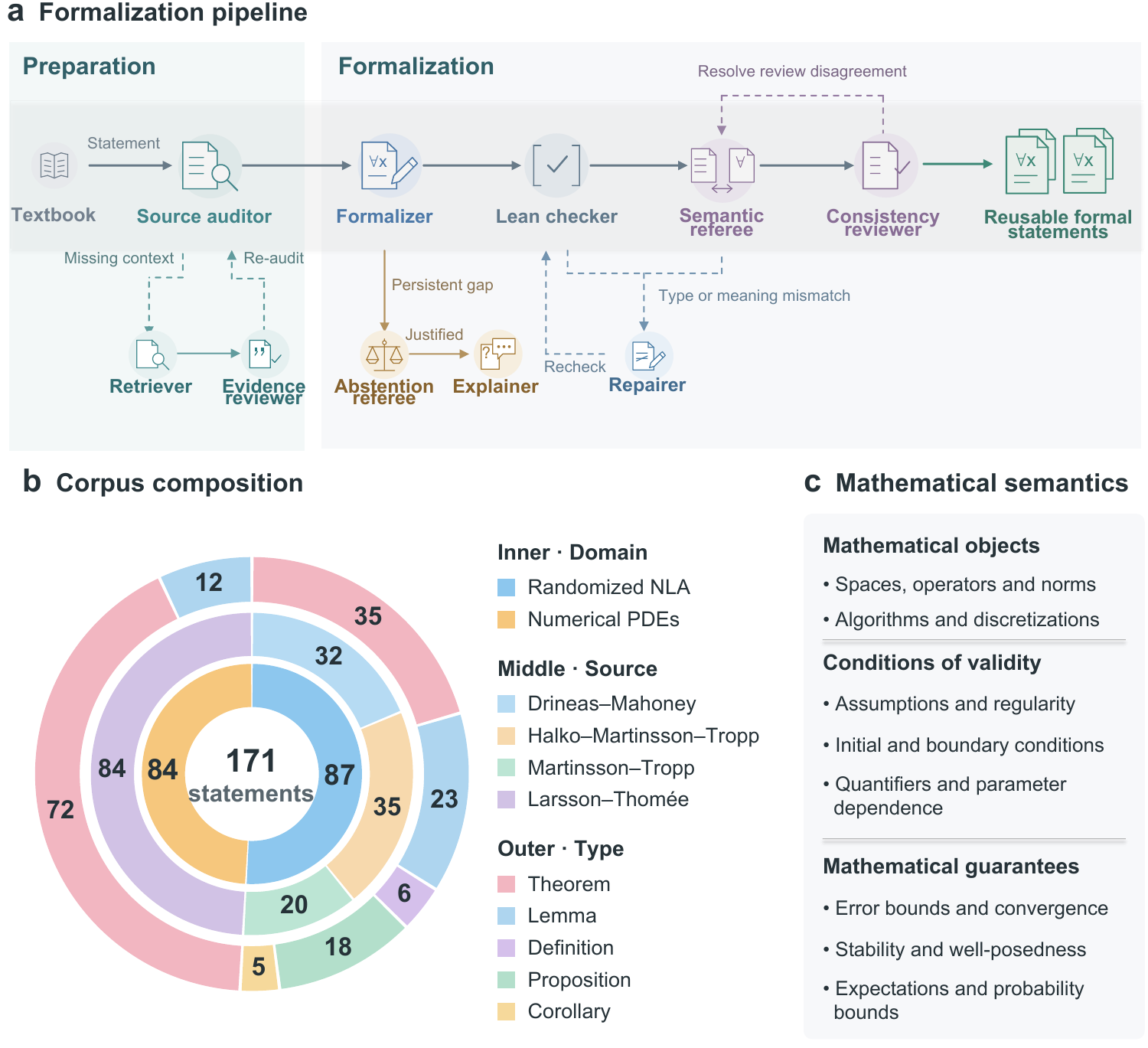}
\caption{\textbf{FARMS builds AI-ready computing knowledge for automated computational science.} \textbf{a}, Source preparation links mathematical statements to supporting definitions, assumptions and algorithm descriptions. Formalization combines this evidence with reusable domain representations, followed by compilation, semantic review and bounded repair. Dashed paths show source re-auditing, repair and rechecking, resolution of review disagreement, and justified abstention. \textbf{b}, The collection comprises 171 targets, shown by domain (inner ring), source (middle) and statement type (outer): 87 from randomized numerical linear algebra and 84 from numerical PDEs. \textbf{c}, Computational structures represented in the collection, including algorithmic constructions and quantitative guarantees.}
\label{fig:framework}
\end{figure}

FARMS builds AI-ready computing knowledge by linking each formal statement to the source material and computational interpretation needed to use it (Figure~\ref{fig:framework}). Each representation makes the mathematical objects, applicability conditions and guarantees explicit, while retaining the evidence behind its interpretation. This organization provides a basis for checking how a computational claim is represented and how it can be connected to other claims.

\paragraph{Preparation.} This stage assembles the source evidence needed to interpret each computational statement (Figure~\ref{fig:framework}a). PDF or LaTeX sources are first converted into text that preserves mathematical expressions, algorithm descriptions and source locations. The Source auditor assesses whether a statement is accompanied by the definitions, assumptions and algorithmic context needed for formalization. When context is missing, the Retriever searches for relevant passages using deterministic retrieval routines guided by the identified gap. Here the target is the source text that fixes a statement's meaning, so the retrieved object is evidence rather than a proof step \cite{yang2023leandojo}. The Evidence reviewer evaluates whether the retrieved material addresses that gap and returns supporting evidence to the Source auditor for re-auditing. This bounded loop links each statement to the source context needed to preserve its computational meaning, including algorithmic steps stated separately from its guarantee.

\paragraph{Formalization.} The Formalizer uses the prepared evidence and shared domain library to construct a Lean statement and record its interpretation of the source. The library supplies reusable representations of mathematical objects and computational constructions, reducing the need to reconstruct common infrastructure for each target. The Lean checker compiles each candidate in the selected formal environment. Compilation establishes that a statement is well formed in the formal environment. The Semantic referee then examines whether the candidate preserves the source's objects, assumptions, quantifiers, algorithmic constructions and quantitative guarantees. Semantic review assesses alignment with the source. A type or meaning mismatch is referred to the Repairer, after which the revised candidate returns for compilation and semantic review. The Consistency reviewer provides a final check on the disposition of findings raised during formalization and review. Review disagreements return to the Semantic referee for resolution before acceptance. 

Statements that meet the acceptance criteria become reusable formal statements, with documented source warnings retained where applicable. A separate branch addresses persistent gaps reported by the Formalizer: the Abstention referee assesses whether abstention is justified, and the Explainer records the reason when it is upheld. These branches distinguish repairable formalization problems from cases in which the available evidence does not support completion.

Figure~\ref{fig:framework}b summarizes the 171-target collection by domain, source and statement type. Figure~\ref{fig:framework}c organizes its computational content into mathematical objects, conditions of validity and mathematical guarantees. Together, source-linked evidence and reusable formal representations make these relationships explicit for subsequent computational use.

\subsection{Preserving computational meaning}

\begin{figure}[!t]
\centering
\includegraphics[width=\linewidth]{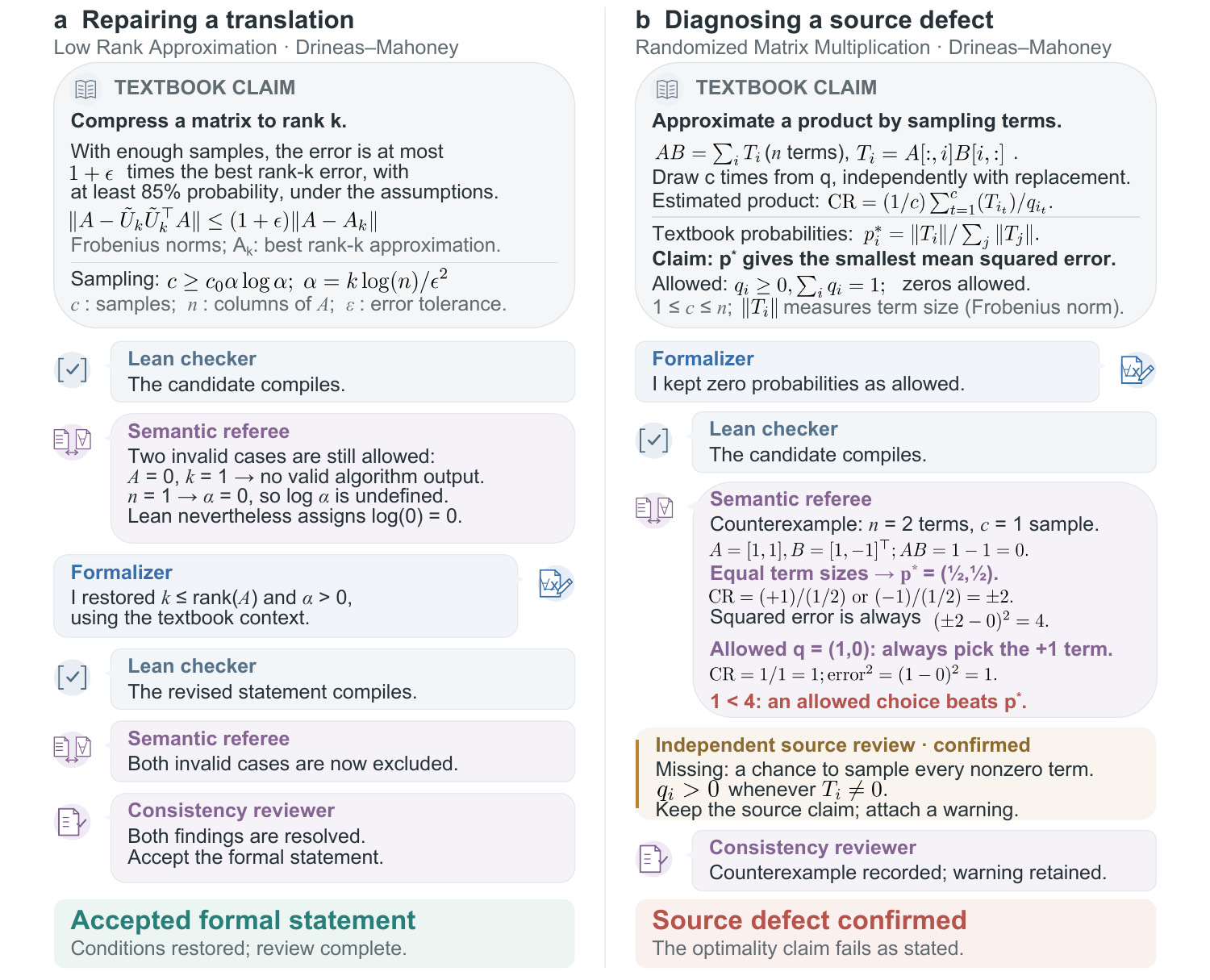}
\caption{\textbf{FARMS refines formalization results and identifies defects in textbook claims.} \textbf{a,} Agent interaction during the formalization of a randomized low-rank approximation guarantee. Although the initial candidate compiles, the Semantic referee identifies omitted rank and logarithm-domain restrictions that change the intended claim. Using evidence from the textbook context, the Formalizer restores these conditions. Subsequent compilation and semantic review confirm the refinement, yielding an accepted formal statement. \textbf{b,} Agent interaction during the examination of the sampling optimality claim in randomized matrix multiplication guarantee. The candidate preserves the probability domain stated in the textbook and passes compilation. The Semantic referee nevertheless constructs a counterexample, and calls an independent source review to confirm a missing condition requiring positive sampling probability for every nonzero contribution, and this finding is kept as an explicit source warning. Together, the examples illustrate how FARMS uses source evidence and mathematical evaluation to distinguish repairable formalization errors from defects in the underlying claims. }
\label{fig:example}
\end{figure}

Figure~\ref{fig:example} illustrates how FARMS coordinates formalization, compilation and semantic review to preserve the mathematical meaning of textbook statements. The two examples follow distinct paths: recovering conditions omitted during formalization and identifying a defect in a source claim.

Figure~\ref{fig:example}a presents the formalization of the randomized low-rank approximation guarantee \cite{drineas2017lectures}. Under the source assumptions, the target requires that, with probability at least $0.85$, a valid output $\tilde U_k$ exists and every valid output satisfies $\|A-\tilde U_k \tilde U_k^{\mathsf T}A\|_F\leq(1+\varepsilon)\|A-A_k\|_F$, where $A_k$ is a best rank-$k$ approximation, $\varepsilon$ is the error tolerance, and $\|\cdot\|_F$ denotes the Frobenius norm. The success event represents an algorithm that can actually produce an admissible approximation. The first message from the Lean checker confirms that the candidate compiles, after which the Semantic referee examines whether its assumptions preserve the intended guarantee. The referee identifies two unintended cases admitted by the candidate. First, $A=0$ and $k=1$ leave no valid output, revealing an omitted rank restriction. Second, when the number of columns of $A$ is $n=1$, the quantity $\alpha=k\log(n)/\varepsilon^2$ vanishes. The source expression $\log(\alpha)$ is then undefined, whereas Lean's definition assigns
$\log(0)=0$. The Formalizer restores $k\leq\operatorname{rank}(A)$ and $\alpha>0$ using the domain of the source formula. The revised statement compiles, the Semantic referee confirms that the intended domain has been recovered, and the Consistency reviewer verifies that both findings have been resolved. With both conditions restored, the statement asserts a guarantee only where the algorithm can produce an admissible output and where the source's own sampling formula is defined.

Figure~\ref{fig:example}b examines the optimality clause of the randomized matrix multiplication guarantee. The source claims that the normalized distribution $p_i^*\propto\|A_{:i}\|_2\|B_{i:}\|_2$ minimizes the expected squared Frobenius error of the sampled approximation $CR$ to $AB$, where $A_{:i}$ and $B_{i:}$ denote the $i$th column of $A$ and row of $B$, respectively. Writing this error as $E(q)=\mathbb{E}_q[\|AB-CR\|_F^2]$, the claim is $E(p^*)\leq E(q)$ for every admissible distribution satisfying $q_i\geq0$ and $\sum_i q_i=1$. The Formalizer preserves this stated domain, including zero probabilities, and the candidate passes compilation. The Semantic referee then constructs an explicit counterexample. For $A=[1,1]$, $B=[1,-1]^{\top}$ and one sample ($c=1$), the exact product is $AB=0$. The norm-based distribution $p^*=(1/2,1/2)$ produces $CR=2$ or $CR=-2$, each with probability $1/2$, giving $E(p^*)=4$. The admissible alternative $q=(1,0)$ produces $CR=1$ deterministically and gives $E(q)=1$. Thus, the claimed optimality would require $4\leq1$. An independent source review is called by the Semantic referee to confirm that the unbiasedness argument requires positive sampling probability for every nonzero contribution, $q_i>0$ whenever $A_{:i}B_{i:}\neq0$, a restriction absent from the source. The pipeline preserves the source claim and records the missing condition as an explicit warning, and the run consequently remains unaccepted.

These examples show why faithful computing knowledge must retain both the intended algorithmic conditions and any unresolved problems in the source. Recovering a missing condition can restore the meaning of a formalization and identifying a source defect makes a limitation explicit before the claim is reused.

\subsection{From faithful statements to scientific computing}

\begin{figure}[p]
  \centering
  \includegraphics[width=\textwidth]{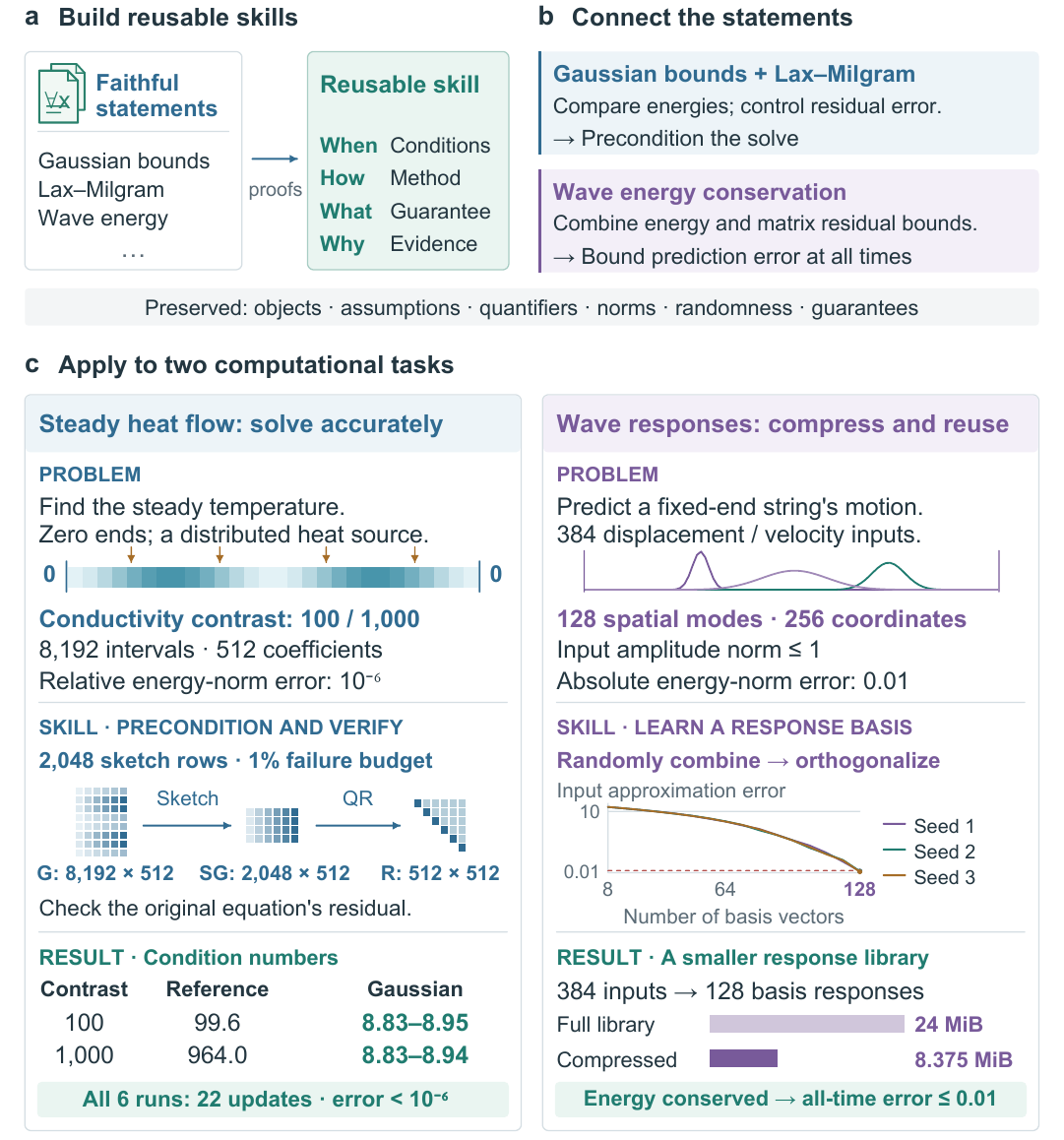}
    \caption{\textbf{From faithful formal statements to numerical agent skills.}
  \textbf{a,} Formal statements and connecting proofs form reusable skills that contain applicability conditions and link results to supporting evidence.
  \textbf{b,} Gaussian bounds and Lax–Milgram support an energy comparison that controls solve accuracy; wave energy conservation and a matrix residual bound support response compression.
  \textbf{c,} Representative applications of the skills. Left, a Gaussian sketch assists the steady heat-flow solve. All six runs meet the $10^{-6}$ relative discrete energy-norm target in a small number of updates. Right, 128 basis responses support wave predictions for 384 prescribed initial disturbances. The response library shrinks from 24 to 8.375 MiB while meeting the 0.01 energy-norm error target. The skills support accurate PDE solving and reusable response models with checkable error bounds.}
  \label{fig:skill}
\end{figure}

To examine how this knowledge can support scientific computation, we connected selected RNLA and PDE statements through Lean-checked implications to form reusable agent skills and used them in two numerical applications (Figure~\ref{fig:skill}). Both applications require an accuracy decision: when an iterative solve is sufficiently accurate, and when a compressed response library can be reused within a prescribed error tolerance.

The connection from statements to skills is shown in Fig.~\ref{fig:skill}a. After establishing the reusable formal statement collection, we assembled existing statements and connecting proofs into procedures that an agent can apply to new cases. Each skill identifies the conditions to check, the method to execute, the guarantee it yields and the evidence supporting that guarantee. Task-specific data and numerical parameters are set at application time, while restrictions of the underlying proofs are fixed. Source statements remain explicit premises: each connecting proof establishes an implication from selected source propositions and application assumptions to a task-specific computational guarantee.

Figure~\ref{fig:skill}b shows the mathematical connections behind these two decisions. The connecting proofs combine statements from our collection with one another or with results already available in Mathlib. Gaussian restricted singular-value bounds establish uniform energy comparisons, while Lax--Milgram and residual estimates connect them to solution accuracy. Wave energy conservation connects an error in the initial state to its error at every later time, while linear algebra supplies rank and residual bounds for compression. The two applications below use the energy-comparison and energy-preservation methods, respectively, with numerical constructions and checks implemented at application time.

The first application solves steady heat flow through a heterogeneous one-dimensional medium with conductivity contrasts of 100 and 1,000 (Figure~\ref{fig:skill}c, left). The discretization has 512 unknown coefficients and a weighted gradient matrix with 8,192 rows. Combining Gaussian singular-value bounds with energy and residual estimates \cite{martinsson2020randomized,larsson2003partial}, the skill selects a 2,048-row sketch to meet a prescribed conditioning requirement under the source and ideal-sampling assumptions. Its QR factor supplies a preconditioner, while the original equation's residual determines when the requested relative discrete energy error of $10^{-6}$ has been reached. Gaussian-preconditioned condition numbers were 8.83--8.95, compared with 99.6 and 964.0 for the constant-conductivity reference preconditioner. All six runs, using three prescribed seeds per contrast, passed the stopping criterion after 22 updates. The sketch thus supplies an auxiliary operator for a checked accuracy decision without reducing the unknown dimension.

The second application compresses a precomputed response library for a string fixed at both ends (Figure~\ref{fig:skill}c, right). The prescribed family contains 384 initial displacement or velocity excitations, represented by an energy-coordinate matrix $F\in\mathbb{R}^{256\times384}$. Queries are combinations $F\alpha$ with $\|\alpha\|_2\leq1$. Gaussian sampling supplies candidate subspaces $Q$ with coefficients $C=Q^{\mathsf T}F$, and a candidate is accepted only when an exact residual check on the saved factors passes. For the ideal wave propagator $W_t$, energy conservation \cite{larsson2003partial} and the matrix residual bound imply $\|W_tF\alpha-(W_tQ)C\alpha\|_E \leq\|F-QC\|_F\|\alpha\|_2.$ Thus one initial-factor check supports an absolute energy-error bound of 0.01 for all allowed queries and nonnegative times, under the stated source premise and exact-model assumptions. All three prescribed seeds selected 128 basis responses, with residual bounds of 0.00877--0.00938. Storing these responses and their combination coefficients reduced the 32-time response library from 24 to 8.375 MiB. This demonstrates accuracy-controlled reuse within a prescribed excitation family.

These applications demonstrate two ways in which formalized computing knowledge can guide a numerical method by forming mathematical skills. In the heat-flow problem, the connected guarantees support parameter selection and an accuracy-based stopping decision. In the wave problem, a bound on the initial representation error combines with energy conservation to control the error of subsequent responses. The collection thus provides reusable mathematical components for reasoning about computations across domains.

\subsection{Quantitative results}

\begin{figure}[p]
  \centering
  \includegraphics[width=0.99\textwidth]{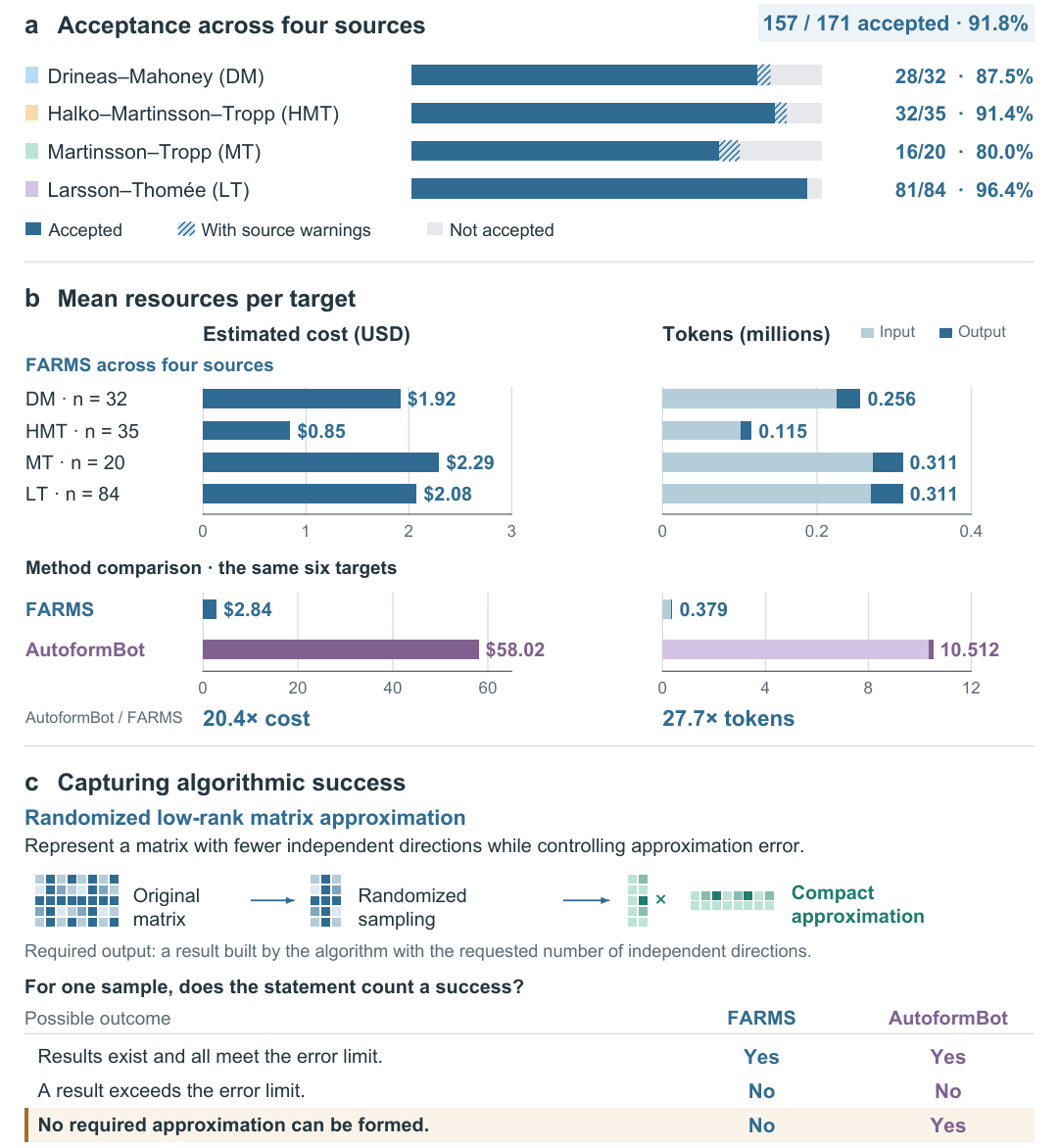}
  \caption{\textbf{Knowledge-construction acceptance, resource use and algorithmic semantics.} \textbf{a,} Cumulative selected outcomes for 171 targets; the 157 reported acceptances include three source-warning results (hatched). \textbf{b,} Mean cost and tokens per planned target, including source preparation, generation, review, repair and continuations within each selected logical run. Source and selected-set plots use separate scales. The six-target comparison uses our statement-level adaptation of AutoformBot. \textbf{c,} Randomized low-rank matrix approximation illustrates FARMS's computing-oriented design. The table contrasts output-existence and accuracy conditions in the two formal statements for the same sample.}
  \label{fig:stat}
\end{figure}

We evaluated knowledge construction on 171 targets from four mathematical sources. The cumulative selected results contained 157 acceptances (91.8\%), comprising 154 without source warnings and three with documented warnings (Figure~\ref{fig:stat}a). A source warning records an unresolved issue in the source claim or its interpretation.

We measured resource use over each selected logical run, including source preparation, generation, review, repair and any continuation needed to complete that run. Mean estimated costs per planned target were \$1.92, \$0.85, \$2.29 and \$2.08 for the four sources, respectively (Figure~\ref{fig:stat}b). The corresponding mean token totals were 0.256, 0.115, 0.311 and 0.311 million. For comparison, we adapted AutoformBot \cite{rammal2026formalizing} to statement-level formalization where we restricted its scope to the selected targets and allowed target and auxiliary theorem proofs to be omitted, while retaining its native source reading, planning, auxiliary-definition construction, compilation, review and repair workflow. We evaluated this adapted version on six randomly selected targets. In total, FARMS used \$17.06 and 2.275 million tokens, whereas AutoformBot used \$348.09 and 63.071 million tokens. These totals correspond to \$2.84 versus \$58.02 and 0.379 versus 10.512 million tokens per target. AutoformBot's recorded cost and token use were therefore 20.4 and 27.7 times those of FARMS in these configurations. 

FARMS is organized around the computational content of mathematical statements. Its source preparation and semantic review connect mathematical guarantees to algorithmic operations, the conditions under which outputs can be constructed, and their accuracy and probability bounds. General-purpose autoformalization workflows pursue different objectives: building textbook-scale libraries with proofs \cite{rammal2026formalizing}, formalizing competition problems at scale \cite{ying2024lean}, or improving terminological and syntactic consistency so that generated declarations fit an existing library \cite{zhang2024consistent}. For the computational guarantee, validity requires both output existence and the stated error bound. This emphasis is related to the explicit constructions evaluated by AMBER and the textbook dependencies recovered by CAM-Bench \cite{yang2026construction,long2026cambench}. Figure~\ref{fig:stat}c illustrates this computing-oriented design through randomized low-rank matrix approximation. The task is to represent a matrix using fewer independent directions while controlling approximation error. The required output is a result constructed by the prescribed algorithm with the requested number of independent directions. For a fixed sample, FARMS's formal statement requires such a result to exist and every allowed result to satisfy the error limit. The compared statement-level AutoformBot formalization retains the error condition but does not require such a result to exist. If no required result can be formed, there is no output that violates the error condition. FARMS explicitly excludes this case from the event counted as successful. The example makes the design emphasis concrete: an algorithmic success guarantee must include the conditions for constructing its output, together with the mathematical error bound. 

Together with the numerical applications, these results support FARMS as a promising, faithful and efficient foundation for automated computational science. Its treatment of computational content makes algorithmic requirements explicit, while its low per-target cost can make the construction and extension of AI-ready computing knowledge more accessible. This combination provides a practical basis for broader community participation in trustworthy automated scientific discovery.

\section{Discussion}
FARMS connects computational mathematics to the decisions required in scientific computing. Its contribution is a reusable form of mathematical knowledge in which methods remain linked to their applicability conditions, quantitative guarantees and source evidence. The heat-flow and wave-response examples show how these links can support parameter selection, accuracy assessment and reuse of numerical results. Shared representations and source-guided review provide an organization for constructing this knowledge, with low per-target costs in the evaluated collection. These results make the expansion of an AI-ready mathematical foundation a practical direction for automated computational science.

The present collection covers four sources in two areas of computational mathematics. Extending its coverage will require handling differences in exposition as well as mathematical content. Missing context and ambiguous formulations remain obstacles, alongside limitations of generation and semantic review. The numerical applications establish that selected statements can support useful computational decisions. Extending these connections to broader problems will require additional mathematical representations and application-specific checks.

A first direction is to integrate knowledge across textbooks. Organizing statements by source preserves provenance, but can leave equivalent results represented several times in different notation or at different levels of
generality. A shared collection should identify these relationships while retaining distinctions in assumptions and guarantees. Formal equivalences and implications could link source-specific formulations to common representations, allowing users to find the version appropriate to a task \cite{gauthier2019aligning}. This would reduce duplication and make results from different areas easier to combine without losing their original context. Related work addresses dependency-aware statement translation \cite{wang2026aria}, diagnosis and correction of semantic failures \cite{wang2026formalrx}, and autoformalization of scientific reasoning \cite{meadows2026formalscience}. These approaches offer complementary ways to extend the range of sources and strengthen the review of computational claims.

A second direction is to improve how the framework adapts to different styles of mathematical exposition. Some sources present computational results through prose, derivations or algorithm descriptions without a separate theorem statement. Feedback from source auditing, compilation and semantic review could help revise extraction and retrieval strategies for such material. Iterative refinement with language-model feedback provides a relevant precedent \cite{madaan2023self}. Faithfulness also requires preserving the distinction between a source claim and a repaired claim. FaithformBench evaluates both validity and invalidity preservation and reports that autoformalizers can silently turn invalid inputs into provable statements \cite{cornish2026faithformbench}. This motivates retaining source warnings and making any proposed correction explicit when constructing reusable mathematical knowledge. The objective is to recover the conditions needed for faithful representation across a wider range of sources while keeping the review effort manageable.

A third direction is to incorporate recent research articles. Textbooks provide established foundations, but new computational methods and sharper guarantees often appear first in the research literature. Retrieval-augmented generation \cite{lewis2020retrieval} offers a way to bring relevant documents into the construction process. Its value here would be access to new statements and supporting context. The retrieved material would still require source auditing, formal representation and review. Linking new results to the conditions and guarantees already in the collection could make advances easier to compare and reuse, while source and version records would support subsequent correction and extension.

The longer-term opportunity is to connect this evolving knowledge base to agents that propose, combine and evaluate computational methods \cite{novikov2025alphaevolve}. The present skills demonstrate useful connections that such agents could draw on. Constructing new connections and discovering new methods remain open research tasks. A community-built collection of faithful statements, checked implications and executable procedures could make accumulated mathematical knowledge a more accessible resource for trustworthy automated scientific discovery.

\section{Methods}
\subsection{Source collection}
We studied statement formalization on 171 targets from four mathematical sources. The randomized numerical linear algebra collection comprised 32 targets from Drineas–Mahoney, 35 from Halko–Martinsson–Tropp and 20 from Martinsson–Tropp. The numerical partial differential equation collection comprised 84 numbered theorems and lemmas from Larsson–Thomée.

\subsection{FARMS workflow}
FARMS first assembled a source package for each target. Preparation preserved the statement and the surrounding definitions, assumptions and algorithm descriptions needed to interpret it. Deterministic retrieval identified candidate source passages, and source auditing and evidence review assessed whether they supplied the required context. Unresolved questions triggered bounded retrieval and re-auditing. Retrieval serves different roles in formal mathematics: LeanDojo retrieves premises for proof generation, and LeanSearch retrieves Mathlib statements from natural-language queries \cite{yang2023leandojo,gao2024semantic}. In FARMS, the Retriever instead recovers source passages that determine a target statement's meaning, including definitions and algorithm descriptions. The package retained the evidence supporting each interpretation, including dependencies between a mathematical guarantee and an algorithm described elsewhere in the source. A shared domain library supplied reusable formal representations of mathematical objects and computational constructions. The formalizer generated Lean declarations together with source-linked interpretations. Lean compilation checked whether the declarations were well formed in the selected environment. A separate semantic review assessed whether the candidate preserved the source objects, assumptions, quantifiers, algorithmic constructions and guarantees. Compilation failures and semantic findings entered bounded repair cycles, followed by renewed checks. Persistent context gaps were referred for source re-auditing and, when unresolved, abstention review. A consistency review checked the disposition of issues raised during the run. We defined acceptance as a final pipeline outcome of either ``VERIFIED\_PASS" or ``COMPILED\_WITH\_SOURCE\_WARNING".

\subsection{Models and execution environments}
The principal model configuration used Anthropic's Claude through the Claude Code command-line interface, with the model selector \texttt{claude-opus-5-5}. Under the role-specific policy, the Source auditor, Formalizer, Semantic referee, Repairer, Consistency reviewer and Abstention referee used \texttt{high} effort. Initial evidence review also used \texttt{high} effort. Context-retrieval planning and quotation-selection calls used \texttt{medium} effort, as did the Explainer. The Retriever combined these planning calls with deterministic document searches and the Lean checker invoked the compiler directly. These tool operations had no language-model effort setting. Our statement-level AutoformBot adaptation used the same Claude selector with \texttt{high} effort.

\subsection{Connecting formal statements to computational skills}
The downstream skills combined selected source propositions with additional Lean-checked implications and numerical procedures. Each skill specified applicability conditions, an executable method and the evidence required to support its conclusion. Source propositions remained explicit mathematical premises of the connecting proofs. The new proofs establish implications from those premises to task-specific guarantees. Dependency checks for the reported connecting proofs excluded ``sorryAx" and retained only the standard logical axioms. The two applications were developed and executed locally using these skills. For each application, we distinguish the mathematical implication checked in Lean, the concrete assumptions supplied by the application, and the numerical checks performed outside Lean.

Reusable formal components already support several forms of mathematical reasoning, including verified convergence guarantees for optimization algorithms \cite{li2025convergence}. LEGO-Prover grows a library of verified lemmas for proof construction \cite{wang2024legoprover}, and Divide and Abstract learns abstractions for statement autoformalization \cite{min2026divide}. FARMS uses shared domain representations to formalize source claims and connects selected claims through Lean-checked implications to numerical accuracy decisions.

\subsubsection{Steady heat-flow application}
Given a distributed source $f(x)=\sin(\pi x)+0.2\sin(5\pi x)+0.1\cos(3\pi x)$ and zero temperature at both ends, the task is to solve \[
-(a_\eta u')'=f\quad\text{on }(0,1),\qquad u(0)=u(1)=0,
\]
with \(a_\eta(x)=\eta+(1-\eta)\sin^2(2\pi x)\), \(\eta\in\{0.01,0.001\}\), giving conductivity contrasts, $\max a_\eta/\min a_\eta$, of 100 and 1,000. The temperature is approximated by 512 trial functions, constructed by linearly interpolating sampled sine modes on 8,192 mesh intervals. Thus, there are 512 unknown coefficients. The mesh creates the gradient and load data. The weighted gradient matrix $G\in\mathbb R^{8192\times512}$ defines the discrete system $Ac=b$, where $A=G^{\mathsf T}G$ is $512\times512$. The requested relative energy accuracy is $10^{-6}$ for this discrete system. The spectral-equivalence skill turns the PDE data into a preconditioning and verification procedure. Using the Gaussian source statement, the application searches candidate sizes against a 1\% failure budget and a requested condition-number upper bound of 30. It selects 2,048 sketch rows, with a sufficient bound of approximately 15.3 under the source and ideal Gaussian sampling assumptions. The resulting sketch compresses the gradient data from 8,192 to 2,048 rows. A QR factorization of $SG$ provides the preconditioner $B=(SG)^{\mathsf T}(SG)$, implemented by triangular solves. The unknown dimension remains 512. Combining the energy comparison with Lax--Milgram and residual estimates gives an accuracy test for the original equation. For contrasts of 100 and 1,000, a diagonal preconditioner from a constant conductivity reference gave condition numbers of 99.6 and 964.0, respectively. Gaussian preconditioning gave condition numbers of 8.83--8.95 across the two contrasts and three prescribed seeds per contrast. In these examples, the sketch provides an auxiliary energy that tracked the heterogeneous operator, maintaining similar preconditioned condition as the conductivity contrast increased. The skill connected this approximate operator to a verifiable accuracy target for the original discrete equation.

\subsubsection{Wave-response compression application}
The displacement satisfies $u_{tt}=u_{xx}$ on $(0,1)$, with zero boundary values. The task specifies localized initial displacement or velocity profiles at 64 positions and three widths, giving 384 excitation channels. Each profile is represented by 128 sine modes. Displacement and velocity  together require 256 state coordinates. In coordinates whose Euclidean norm equals the physical energy norm, the columns of $F\in\mathbb R^{256\times384}$ describe these initial states. A query gives amplitudes $\alpha$ with $\|\alpha\|_2\le1$, and asks for the complete response to $F\alpha$. The target is an absolute energy-norm error of 0.01 over this amplitude class. The isometry skill makes this a problem of approximating the excitation map once. The application forms Gaussian combinations of its columns, orthogonalizes them and adds eight basis directions at a time. For each candidate basis $Q$, the coefficients $C=Q^{\mathsf T}F$ express the original excitations in that basis. The procedure accepts a factorization only after its residual meets the target, including an exact check of the saved factors. If $W_t$ denotes wave propagation, the resulting prediction is $(W_tQ)C\alpha$. The application propagates the basis states and reuses their responses for different queries. Wave energy conservation and the matrix residual bound ensure that checking the initial factorization supports every later time. All three prescribed seeds selected 128 basis directions, with checked Frobenius residuals of approximately 0.00877--0.00938. Each selected model was evaluated on the 384 individual excitations, 128 random amplitude combinations and a worst residual direction at six times. Replacing 384 excitation responses with 128 basis responses reduced float64 array storage from 24 to 8.375\,MiB. The outcome is a smaller response library with an error budget that applies to all allowed queries and times. In this example, the learned basis provided a shared representation of the prescribed excitation family, reducing the storage required for its response library. The skill connected this compact representation to a verifiable error bound that remained valid for every allowed excitation combination throughout wave evolution.

\subsection{Statement-level AutoformBot comparison and resource accounting}
We compared FARMS with a statement-level adaptation of AutoformBot \cite{rammal2026formalizing} on selected RNLA targets. Starting from the archived upstream revision, we restricted the task to the supplied target list and permitted `by sorry` in target and auxiliary theorem proofs. The adaptation retained the native source reading, retrieval, planning, auxiliary-definition construction, compilation, review, repair, retry and merge workflow, with the execution and model-interface changes required by our environment. Definitions retained the original requirements. AutoformBot received the same complete source documents. For each target, we measured the resources of the selected logical run, including source preparation, generation, review, repair and recorded continuations. Mean resources were computed by dividing the total by the number of planned targets, including targets that were not accepted.



\section{Acknowledgements}
H.Y. was partially supported by the US National Science Foundation under awards IIS2520978, GEO/RISE-5239902, the DARPA D24AP00325, and the 2026 AIM Seed Grant from the AI Interdisciplinary Institute at Maryland (AIM). Approved for public release; distribution is unlimited.





\bibliographystyle{unsrtnat}
\bibliography{references}

\end{document}